\documentclass{article}
\usepackage{arxiv}
\usepackage{listings}

\usepackage{caption}
\usepackage{graphicx}        
\usepackage{fixltx2e}
\usepackage[section]{placeins}
\usepackage{float}
\usepackage[utf8]{inputenc}

\usepackage[usestackEOL]{stackengine}

\usepackage{amsmath}
\usepackage{array}
\usepackage[final]{pdfpages}
\usepackage[ampersand]{easylist}
\usepackage{color}
\usepackage[overload]{textcase} 

\usepackage{booktabs}

\usepackage{listings}
\begin{document}

\title{Extracting Aspects Hierarchies using Rhetorical Structure Theory}

\author{
    Łukasz Augustyniak, Tomasz Kajdanowicz and Przemysław Kazienko \\
    Wrocław University of Science and Technology, 
    Wrocław, Poland \\
    lukasz.augustyniak@pwr.edu.pl, 
    tomasz.kajdanowicz@pwr.edu.pl, 
    przemyslaw.kazienko@pwr.edu.pl
    }

\maketitle

\begin{abstract}
We propose a~novel approach to generate aspect hierarchies that proved to be consistently correct compared with human-generated hierarchies. We present an unsupervised technique using Rhetorical Structure Theory and graph analysis. We evaluated our approach based on 100,000 reviews from Amazon and achieved an astonishing 80\% coverage compared with human-generated hierarchies coded in~ConceptNet. The method could be easily extended with a~sentiment analysis model and used to describe sentiment on different levels of aspect granularity. Hence, besides the flat aspect structure, we can differentiate between aspects and describe if the \textit{charging} aspect is related to \textit{battery} or \textit{price}.

\end{abstract}

\section{I\MakeLowercase{ntroduction}}
\label{sec:Introduction}

Aspect-based sentiment analysis deals with classifying sentiment towards a~given aspect in~documents. Most of the time sentiment classification approaches work on whole documents, and they do not distinguish between the various aspects mentioned in~the text. Imagine that you have a~sentence such as \textit{the wall charger works, but the car one does not}. This example expresses two attitudes toward two different aspects of the phone charging process. Firstly, we spot positivity for the \textit{wall charger}, and secondly negativity for the \textit{car charger}. Such an analysis is not feasible with standard sentiment analysis tools. We used the discourse analysis as the text segmentation tool and extracted with it the basic units called Elementary Discourse Units (EDUs). Moreover, the discourse analysis extracts the connection between EDUs and determine its semantic relations. We conduct analysis separately for each review to get Discourse Trees (DTs) such as in~Figure~\ref{fig:ra-example}. Beside the flat aspect structure sometimes we want to know how different aspects are related to each other.  For example, if the \textit{charging} is related to \textit{battery} or price \textit{aspect}? This approach needs relation or hierarchies between aspects to be derived. To deal with such problems we need hierarchical aspect-based sentiment models. 

\begin{figure}[!ht]
\centering
\includegraphics[width=\columnwidth]{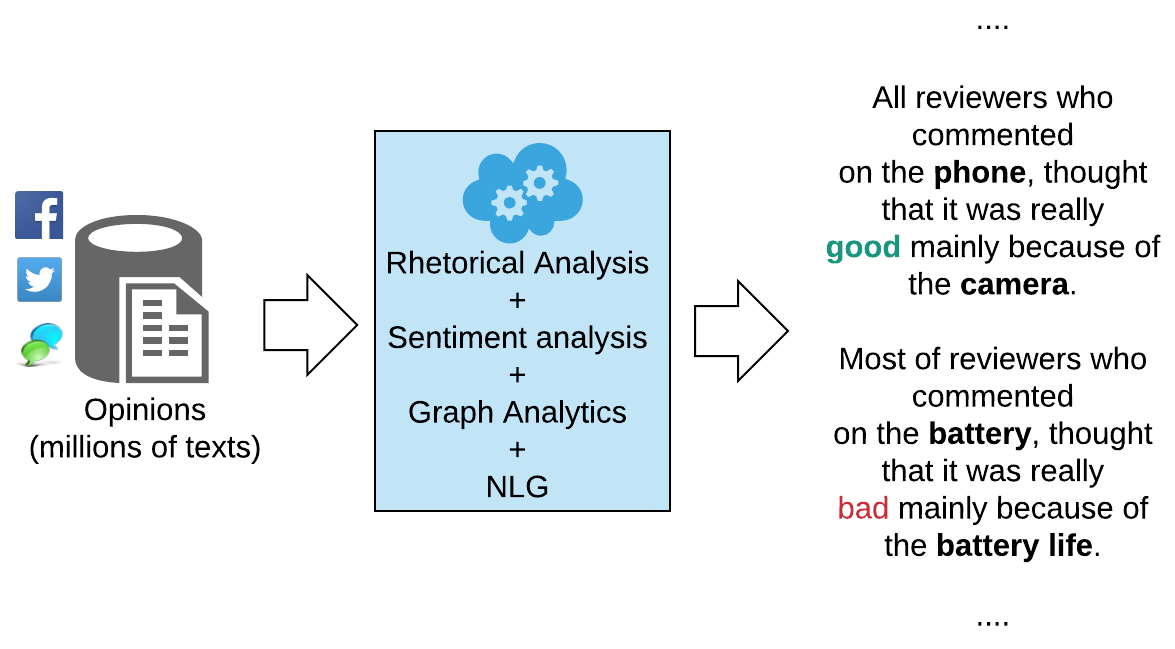}
\caption{The Workflow for Sentiment Analysis Summarization using Rhetorical Structure Theory.\label{fig:introflow}}
\end{figure}

We are working on a~comprehensive pipeline for analysing texts containing opinions and generating user-friendly descriptive, abstractive reports in~natural language (Figure~\ref{fig:introflow}). In~this paper, we describe extending our previous work \cite{Augustyniak2017} and present a~prototype implementation of aspect hierarchy extraction.

The paper presents in~Section~\ref{sec:Introduction} an introduction to hierarchical aspect-based sentiment analysis. Then in~Section~\ref{sec:RelatedWork} we present related work from the areas of rhetorical and aspect-based sentiment analysis. In the Section~\ref{sec:TheProposedMethod} we described our method.Then the implementation and the data are described in~Section~\ref{sec:ProblemDescription}. The Section~\ref{sec:Results} contains the results and thier description. The last Section~\ref{sec:Conclusions} covers our conclusions and future work.

\section{R\MakeLowercase{elated} W\MakeLowercase{ork}}
\label{sec:RelatedWork}

\subsection{Rhetorical Structure Theory}

Rhetorical Structure Analysis (RSA) tries to uncover the coherence structure of the texts. This technique has been shown to be beneficial for many Natural Language Processing (NLP) applications such as sentiment analysis \cite{Lazaridou2013}, text summarization \cite{Louis2010} or machine translation \cite{Guzman2014}. There exist various theories for RSA, such as Martin \cite{Martin1992} and discourse relations based on discourse connectives, words such as \textit{because} or \textit{but}. Beside that, Danlos \cite{Danlos2011} extended sentence grammar and formalized discourse structure analysis. 

\begin{figure}[!ht]
\centering
\includegraphics[width=\columnwidth]{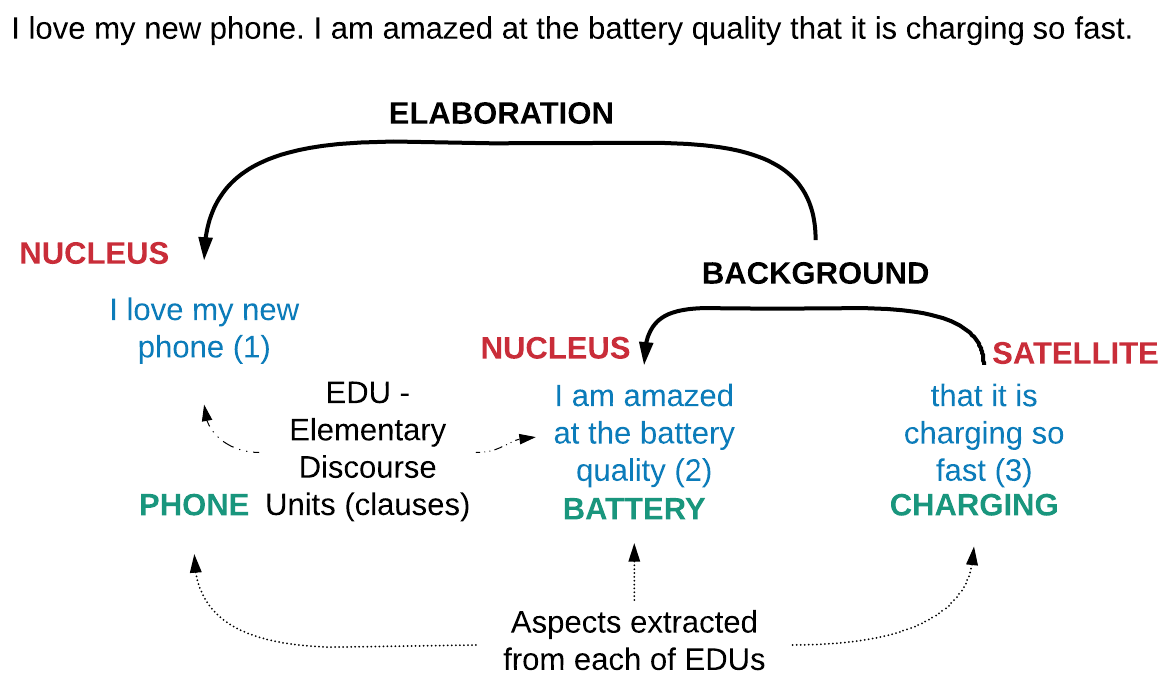}
\caption{An Exemplary Discourse Tree Based on Rhetorical Structure Theory.}
\label{fig:ra-example}
\end{figure}

However, we use the Rhetorical Structure Theory (RST) by Mann and Thompson \cite{Mann1988}. This method is probably the most influential theory of discourse analysis in~computational linguistics. It was constructed especially for the text generation tasks. Then it became popular for text parsing \cite{Taboada2006}. Rhetorical Structure Theory represents documents as tree structures with relations between different parts of the text. Based on RST we can generate tree structures called Discourse Trees (DTs). An exemplification of a~DT is presented at Figure~\ref{fig:ra-example}.

\subsection{A\MakeLowercase{spect hierarchies in}~S\MakeLowercase{entiment} A\MakeLowercase{nalysis}}

As far as we know, there are only a~couple of works on the hierarchy extraction of aspects for sentiment analysis purposes. Kim et al. \cite{Kim2010} proposed a~hierarchical aspect sentiment model (HASM) and tried to discover a~hierarchical structure of aspects from unlabeled documents. The HASM represents the whole structure as a~tree. Secondly, Wei and Gulla \cite{Wei2010} proposed a~Sentiment Ontology Tree, but unfortunately, they analyzed the reviews of one product by manually labeling the product’s aspects with sentiments.

Researchers use several approaches to aspect-based sentiment analysis. From still commonly used rule-based methods (POS \cite{Augustyniak2017, Poria2016} or dependency-based \cite{Poria2014}), through standard supervised learning (e.g., SVMs and CRF \cite{Toh2014}) to deep learning-based approaches with CNNs or LSTMs. There is an interesting approach proposed by Ruder et al. \cite{Ruder2016}. He used a~hierarchical, bidirectional LSTM model to leverage both intra and inter-sentence relations. Poria et al. \cite{Poria2016} proposed a~seven-layer convolutional neural network to tag each word in~opinionated sentences as either an aspect or non-aspect word. However, these approaches only extract a~flat structure of aspects from the text.

There exist several models for discovering topic hierarchies from textual data such as \cite{Blei:2010:NCR:1667053.1667056, Kim:2012:MTH:2396761.2396861}, but they are not designed for aspect-based hierarchies.

\section{M\MakeLowercase{ethod for} a\MakeLowercase{spect} h\MakeLowercase{ierarchies} g\MakeLowercase{eneration}}
\label{sec:TheProposedMethod}

Our hierarchy extraction model consists of the following sub-tasks.

\begin{enumerate}
\item Rhetorical analysis of documents, segmentation of the text into the basic units of discourse structures: EDUs (Elementary Discourse Units).
\item Aspect detection for each of the EDUs, creating Aspect Discourse Trees (ADTs).
\item Hierarchy generation based on RST relations and extracted aspects.
\end{enumerate}

We hypothesize that when aggregating aspects and the rhetorical relation between these aspects for many documents we get aspect hierarchical relationships such as that the \textit{phone} is a~precedent for the \textit{battery} and the \textit{battery} is precedent for \textit{battery life}. Why do we need a~lot of documents? Each separate document and its aspects can be rather noisy because of (1) the low level of formality of the language of the reviews and (2) inaccuracies from automatic discourse segmentation tools. Hence, by   aggregating a~lot of documents we may spot more reliable and general information. It is worth mentioning that in~the most similar approach to ours, Joty et al. \cite{Joty2015} extracted only single aspect relations from every document. Hence, they may potentially miss some of the informative value.

\subsection{Rhetorical Analysis and Aspect Hierarchy Generation}

We used the discourse analysis as the text segmentation tool and extracted the basic units called Elementary Discourse Units (EDUs). Moreover, the discourse analysis extracts the connection between EDUs and determine its semantic relations. We conduct analysis separately for each review to get Discourse Trees (DTs) such as in~Figure~\ref{fig:ra-example}. 

\subsection{Aspect detection in~textual data}

\begin{figure}[!ht]
\centering
\includegraphics[scale=0.25]{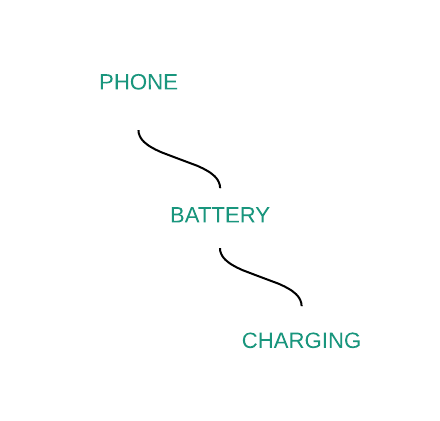}
\caption{An exemplary Aspect Hierarchical Relation extracted from a~single Aspect Discourse Tree.\label{fig:ra-example-aspects}}
\end{figure}

Then we extracted aspects and created aspect-based discourse trees (ADTs). Aspect detection from textual data is based commonly on detection of names or noun-phrases \cite{MariaPontikiDimitriosGalanisHarisPapageorgiouSureshManandhar2015} and we used exactly this approach. We replace EDUs with aspects extracted from them, while also skipping all nucleus-nucleus relations (these relations do not show the hierarchical dependency between aspects). At the end of this step, we get trees as in~Figure~\ref{fig:ra-example-aspects}.

\subsection{Aspect Hierarchy generation}
Then we gather relations from all Aspect Discourse Trees and extract general hierarchical relations between aspects. We generate tuples from ADTs $(a_{i}, a_{j}, r)$, $a_{i}$ and $a_{j}$ mean aspects, $i$ and $j$ indicate the aspect's index and $i \neq j $, $r$ state the rhetorical relations type. We use a~breadth-first search (BFS) \cite{Eppstein2009} algorithm for each Discourse Tree to generate the tuples mentioned. These tuples can be simplified to $(a_{i}, a_{j})$ on the assumption that the first element of tuple $a_{i}$ is nucleus and the second element $a_{j}$ means a~satellite.As a~reminder, the nucleus means the more important part of the RST relation and a~satellite is a~subsidiary part. We count all such tuples and use the top $n$ aspects appearing in~these tuples, as in~Listing~\ref{lst:ns}. We start with the most common tuple \textit{(phone, case)}, where \textit{phone} is the nucleus (hence it is the root node in~the hierarchy), and the satellite in~this pair \textit{case} will be the first child in~the hierarchy for the root node. Then, in~the second step, we go to the next element in~the sorted pairs \textit{(phone, battery)}. The nucleus \textit{phone} is already in~the hierarchy hence we will add its child only, \textit{battery}.

\begin{lstlisting}[frame=single,caption={Aspects pairs of Nucleus and Satellite},captionpos=b,label={lst:ns}]
(NUCLEUS, SATELLITE)
(phone, case)
(phone, battery)
(phone, headset)
(phone, bluetooth)
(price, battery)
(battery, charge)
(headset, bluetooth)
(case, phone)
(phone, price)
...
\end{lstlisting}

If we find a~nucleus aspect that is not already present in~the hierarchy such as \textit{(price, battery)} we create a~separate hierarchy with the nucleus as a~root node of this sub-hierarchy. This subtree could be merged to the main hierarchy in~the next steps when in~our pair list it will appear tuple with satellite equal to the root node of this sub-hierarchy, and our main hierarchy will contain a~nucleus from this aspect’s pair such as \textit{(phone, price)}. We repeat the step by going over and over pairs of nucleus and satellite until the end of the list.

It is important to mention that we can add repeated aspects to the hierarchy as children. For example, the pairs \textit{(phone, bluetooth)} and \textit{(headset, bluetooth)} contain the same satellite \textit{bluetooth}. This is a~very interesting phenomenon related to the local structures of the aspect’s sub-hierarchies (see Section~\ref{sec:Results}).

\section{E\MakeLowercase{xperimental} S\MakeLowercase{cenario}}
\label{sec:ProblemDescription}

We used \cite{Feng2014} for Rhetorical Segmentation in~our experiments.We used a~noun and noun phrase extractor according to the part-of-speech tagger from the Spacy Python library. We implemented our own version of the breadth-first search (BFS) algorithm.

\subsection{Dataset}
\label{sec:dataset}

We used a~dataset of Amazon product reviews scraped by Julian McAuley \cite{He2016}. The dataset consists of review's texts, product metadata, and links written between May 1996 and July 2014. We chose one of the domains of this datasets, to be specific Cell Phones and Accessories, and we sampled randomly 100,000 reviews out of 3,447,249 reviews in~total. 

\begin{table}
\caption{Amazon Cell Phones and Accessories dataset statistics.}
\label{tab:dataset}
\begin{minipage}{\columnwidth}
\begin{center}
\begin{tabular}{ll}
\toprule
Metric & Value \\
\midrule
Total \# of reviews & 100,000 \\
Average \# of words in~review & 97 \\
Average \# of sentences in~review & 5.27 \\
\# of reviews with at least 2 aspects  & 98,528 \\
\# of reviews with at least 10 aspects  & 35,219 \\
\bottomrule
\end{tabular}
\end{center}
\end{minipage}
\end{table}

As we can see in Table \ref{tab:dataset}, Amazon reviews are rather lengthy, and most of the time they contain more than two aspects. Moreover, several thousands of reviews consist of more than ten aspects. Hence, we want to extract more than just one aspect pair from each review, in~contrast to \cite{Gerani2014}. 

\begin{figure*}[ht]
\includegraphics[width=\textwidth]{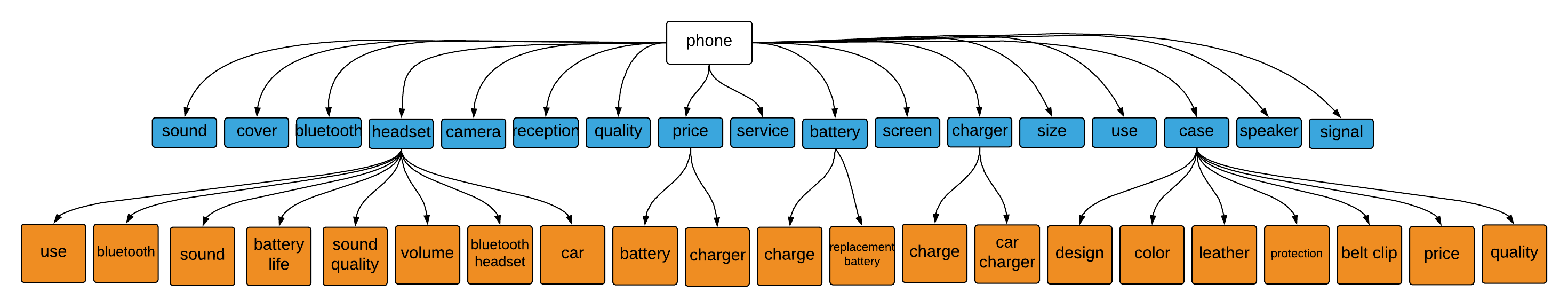}
\caption{Aspect Hierarchy extracted from Amazon Cell Phones and Accessories dataset.\label{fig:aspects-hierarchy}}
\end{figure*}

\subsection{Evaluation}

This section describes the evaluation metrics of our method. A~comprehensive comparison of our method is difficult. To best of our knowledge, there is no prior work where an aspect hierarchy has been described that could easily be compared with our method. Hence, we used a~hierarchical relation expressed in~ConcepetNet \cite{Speer}. ConceptNet is a~graph-based knowledge representation. The ConceptNet contains words and common phrases (both called as concepts) in~any written human language. These words and phrases are related via several different types of relations. These relations describe not only lexical definitions but also how concepts are related based on common knowledge. A~couple of examples are \textit{keypad PartOf phone}, \textit{IsA tool for take picture}, \textit{camera IsA device}, \textit{flash PartOf camera}, \textit{camera IsA photo device}, \textit{car MadeOf metal}, \textit{car HasA seats}, \textit{car HasA windows} and others. 

Importantly ConceptNet covers relations of a~hierarchical nature such as presented by Mukherjee and Joshi in~\cite{Mukherjee2013} as in~\textit{LocatedNear, HasA, PartOf, MadeOf and IsA}.

Moreover, ConceptNet also covers relation such as \textit{Synonym} and \textit{RelatedTo}. These two relations were very useful and enabled us to obtain hierarchical relations not expressed directly in~ConceptNet. For example, we could find the relation \textit{receiver PartOf telephone}, but there isn't the relation \textit{receiver PartOf phone}, although using \textit{telephone Synonym phone} relation we can derive \textit{receiver PartOf phone}.

We used the coverage metric to evaluate our method

\begin{equation}
coverage = \frac{|A \cap C|}{|A|}
\end{equation}

\noindent where $A$ is a set of aspect hierarchical tuples generated by our method, and $C$ is a set of all hierarchical tuples extracted from ConceptNet within a~graph distance up to three hops. Reporting the results with the coverage we show of how many of aspect hierarchical tuples discovered by our method exist in ConceptNet. As an example, the relation between \textit{receiver} and \textit{telephone} is equal to 1, because of the \textit{receiver PartOf telephone} explicit relation, but the distance for \textit{antenna} and \textit{telephone} will be 2, because of the relations \textit{antenna PartOf receiver} and \textit{receiver PartOf telephone}. We chose a~distance of up to three because ConceptNet is a~crowd-sourced database and it misses a~lot of relations that should appear in~a~complete solution. With a~distance up to three we spot reasonable hierarchical relations such as \textit{phone} and \textit{case}, \textit{phone} and \textit{battery}, and others.

\section{R\MakeLowercase{esults}}
\label{sec:Results}

In this section, we describe and analyze the results of our method.

We designed our model to produce a~hierarchical structure of aspects such that this hierarchy can be traversed to find certain aspects. Moreover, it can be easily extended with a~sentiment analysis module to enrich aspects with sentiment orientation distribution. The hierarchy of aspects is consistent with our intuition and the root node is the most general aspect/type of the product or service expressed in~all documents. With the depth of hierarchy, the aspects become more and more specific features.

\begin{table}[ht!]
\caption{Coverage values between our Aspect Hierarchical Pairs and ConceptNet Hierarchical Relations.}
\label{tab:coverage}
\begin{minipage}{\columnwidth}
\begin{center}
\begin{tabular}{ll}
\toprule
&  Coverage \\
\midrule
Top 5  &             1.00 \\
Top 10 &             0.90 \\
Top 20 &             0.78 \\
Top 30 &             0.75 \\
Top 40 &             0.82 \\
Top 50 &             0.80 \\
\bottomrule
\end{tabular}
\end{center}
\end{minipage}
\end{table}

Table~\ref{tab:coverage} presents results of the coverage calculated for a~different number of most frequent tuple hierarchical aspect pairs generated by our method. We tested the top 5, 10, 20, 30, 40, and 50 pairs of aspects. We consistently received a~coverage level higher than 75\% for every subset of our hierarchical aspect pairs. This proves that our unsupervised method generates aspect hierarchies consistent with human-generated relations coded in~ConceptNet. Interestingly, all our top 5 aspect pairs were confirmed by ConceptNet relations. Next, the top 10 pairs proved to be 90\% correct which is an astonishing result. Moreover, even the top 30, 40 and 50 pairs achieve a~coverage level of about 80\%.

Interestingly, we can spot the same aspects in~different parts of the hierarchy (see Figure~\ref{fig:aspects-hierarchy}). This is related to the contextuality of aspects. We can have aspects such as \textit{battery} or \textit{bluetooth} in~several sub-hierarchies and each time this aspect will carry different semantics: on the one hand, the battery could be a~battery of the whole phone, on the other hand it could be a~battery of the headset. As we can see, aspects should not be unique in~the hierarchy and they can carry very interesting insights from sentiment analysis tasks.

\section{C\MakeLowercase{onclusions and} \MakeLowercase{Future} W\MakeLowercase{ork}}
\label{sec:Conclusions}

We have proposed a~novel approach to generate aspect hierarchies that proved to be 80\% correct, consistent with the same accuracy levels of human generated hierarchies. The advantages of our method are firstly that it is not limited to the number of aspects, secondly and really importantly, it doesn't need training data (an unsupervised method), and thirdly it is simple to calculate and it could easily be extended with a~sentiment analysis model to provide sentiments on different levels of aspect granularity. in~future work, we want to improve the aspect extraction phase using sequence tagging approaches with BiLSTM and CRF models. We want to extend our method with other metrics to rank aspect pairs. Moreover, we want to apply the analysis to the Polish language.

\section*{Acknowledgment}

The work was partially supported by the National Science Centre, Poland, grant number DEC2016/21/N/ST6/02366, from the European Union's Horizon 2020 research and innovation programme under the Marie Skłodowska-Curie grant agreement No. 691152 (RENOIR) and the Polish Ministry of Science and Higher Education fund for supporting internationally co-financed projects in~2016-2019 (agreement no. 3628/H2020/2016/2), and by the Faculty of Computer Science and Management, Wrocław University of Science and Technology statutory funds.


\bibliographystyle{unsrt}
\bibliographystyle{ACM-Reference-Format}
\bibliography{mlnlp}  
\end{document}